\documentclass[twocolumn]{bmcart}% uncomment this for twocolumn layout and comment line below
\usepackage[utf8]{inputenc} %unicode support
\usepackage{caption}
\usepackage{amsmath,graphicx,amsfonts,epsfig,epstopdf,datetime,url}
\usepackage{booktabs}

\startlocaldefs
\endlocaldefs

\begin{document}

%%% Start of article front matter
\begin{frontmatter}

\begin{fmbox}
%\dochead{CSLT TECHNICAL REPORT-20150016 [\today]}
%\dochead{RESEARCH}


\title{THCHS-30 : A Free Chinese Speech Corpus}

%%%%%%%%%%%%%%%%%%%%%%%%%%%%%%%%%%%%%%%%%%%%%%
%%                                          %%
%% Enter the authors here                   %%
%%                                          %%
%% Specify information, if available,       %%
%% in the form:                             %%
%%   <key>={<id1>,<id2>}                    %%
%%   <key>=                                 %%
%% Comment or delete the keys which are     %%
%% not used. Repeat \author command as much %%
%% as required.                             %%
%%                                          %%
%%%%%%%%%%%%%%%%%%%%%%%%%%%%%%%%%%%%%%%%%%%%%%

\author[
   addressref={aff1},                   % id's of addresses, e.g. {aff1,aff2}
%   noteref={n1},                             % id's of article notes, if any
%   corref={aff1},                                  % id's of article notes, if any
   email={wangdong99@mails.tsinghua.edu.cn}  % email address
]{\inits{DW}\fnm{Dong} \snm{Wang}}
\author[
   addressref={aff1},                   % id's of addresses, e.g. {aff1,aff2}
%   noteref={n1},                             % id's of article notes, if any
   email={zxw@cslt.riit.tsinghua.edu.cn}  % email address
]{\inits{XWZ}\fnm{Xuewei} \snm{Zhang}}

%\author[
%   addressref={aff1,aff3},                   % id's of addresses, e.g. {aff1,aff2}
%%   noteref={n1},
%   corref={aff1},                                  % id's of article notes, if any
%   email={liuc@cslt.riit.tsinghua.edu.cn}  % email address
%]{\inits{CL}\fnm{Chao} \snm{Liu}}

\address[id=aff1]{%                           % unique id
  \orgname{Center for Speech and Language Technology, Research Institute of Information Technology, Tsinghua University}, % university, etc
  \street{ROOM 1-303, BLDG FIT},                     %
  \postcode{100084}                                  % post or zip code
  \city{Beijing},                                    % city
  \cny{China}                                        % country
}

\begin{artnotes}
%\note{Sample of title note}     % note to the article
%\note[id=n1]{Equal contributor} % note, connected to author
\end{artnotes}

\end{fmbox}% comment this for two column layout

%%%%%%%%%%%%%%%%%%%%%%%%%%%%%%%%%%%%%%%%%%%%%%
%%                                          %%
%% The Abstract begins here                 %%
%%                                          %%
%% Please refer to the Instructions for     %%
%% authors on http://www.biomedcentral.com  %%
%% and include the section headings         %%
%% accordingly for your article type.       %%
%%                                          %%
%%%%%%%%%%%%%%%%%%%%%%%%%%%%%%%%%%%%%%%%%%%%%%

\begin{abstractbox}

\begin{abstract} % abstract

Speech data is crucially important for speech recognition research. There are quite some speech databases that can be purchased at
prices that are reasonable for most research institutes. However, for young people who just start research activities or those
who just gain initial interest in this direction, the cost for data is still an annoying barrier.
We support the `free data' movement in speech recognition: research institutes (particularly supported by public funds) publish their data
freely so that new
researchers can obtain sufficient data to kick off their career.
In this paper, we follow this trend and release a free Chinese speech database THCHS-30 that can be used
to build a full-fledged Chinese speech recognition system. We report the baseline system established with this database, including the performance
under highly noisy conditions.

%\parttitle{First part title} %if any
%Text for this section.

%\parttitle{Second part title} %if any
%Text for this section.
\end{abstract}

%%%%%%%%%%%%%%%%%%%%%%%%%%%%%%%%%%%%%%%%%%%%%%
%%                                          %%
%% The keywords begin here                  %%
%%                                          %%
%% Put each keyword in separate \kwd{}.     %%
%%                                          %%
%%%%%%%%%%%%%%%%%%%%%%%%%%%%%%%%%%%%%%%%%%%%%%


% MSC classifications codes, if any
%\begin{keyword}[class=AMS]
%\kwd[Primary ]{}
%\kwd{}
%\kwd[; secondary ]{}
%\end{keyword}

\end{abstractbox}
%
%\end{fmbox}% uncomment this for twcolumn layout

\end{frontmatter}

%%%%%%%%%%%%%%%%%%%%%%%%%%%%%%%%%%%%%%%%%%%%%%
%%                                          %%
%% The Main Body begins here                %%
%%                                          %%
%% Please refer to the instructions for     %%
%% authors on:                              %%
%% http://www.biomedcentral.com/info/authors%%
%% and include the section headings         %%
%% accordingly for your article type.       %%
%%                                          %%
%% See the Results and Discussion section   %%
%% for details on how to create sub-sections%%
%%                                          %%
%% use \cite{...} to cite references        %%
%%  \cite{koon} and                         %%
%%  \cite{oreg,khar,zvai,xjon,schn,pond}    %%
%%  \nocite{smith,marg,hunn,advi,koha,mouse}%%
%%                                          %%
%%%%%%%%%%%%%%%%%%%%%%%%%%%%%%%%%%%%%%%%%%%%%%

%%%%%%%%%%%%%%%%%%%%%%%%% start of article main body
% <put your article body there>

%%%%%%%%%%%%%%%%
%% Background %%
%%

\section{Introduction}
\label{sec:intro}

\subsection{Speech recognition and data}

Automatic speech recognition (ASR) has been an active research area for several decades. Traditional ASR systems are based on the hidden Markov model (HMM)
to represent temporal dynamics of speech signals and the Gaussian mixture model (GMM) to represent distributions of the signals within a
stationary short period of time that generally corresponds to a pronunciation unit, e.g., a phone. This HMM-GMM paradigm dominates ASR research for nearly three decades,
until a few years ago when deep learning stepped in and attained revolutionary success. Although several variants have been proposed, the
most popular deep learning approach for ASR is the so called HMM-DNN hybrid architecture, where the HMM framework is retained while the
GMM is substituted for a deep neural network (DNN) to model static (stationary) distributions of speech signals. Readers can refer to
~\cite{RECENT,Perspective,automatic,Deep,Fast,Discriminative} for details about DNN-based ASR methods.

Due to the great improvement caused by DNN and other factors (e.g., mobile network and big data), speech
recognition has gained much attention in the research community and multiple industrial sectors. On one hand, the greatly
improved accuracy (in terms of word error rate, WER) has enabled broad practical applications, e.g., voice search on mobile devices; on the
other hand, the technique is still fragile and it is not easy to recognize speech with strong accents and in complex noise conditions. It is also
not simple to migrate a system that works well in one domain to another if the difference is significant. This subtle situation attracts much interest
on speech technologies among researchers, both young graduates and senior researchers who previously work on different areas. This is certainly good news
since more researchers mean more rapid advance in this field, and more close we are from the ultimate goal of understanding human speech.

Nevertheless, there is a big barrier that is hindering researchers joining the ASR community: data. Everybody knows that ASR research heavily
relies on data, even more heavily than other areas that look similar, e.g., face recognition. This is because human speech is so complex that the essential
voice patterns can be learned only with a large amount of data.

\subsection{Data is blocking research}

There are indeed quite some speech databases available for ASR research, e.g., the TIMIT database~\cite{Phoneme} that was designed for
small-scale phone recognition, the WSJ database~\cite{Design} that was designed for large-scale continuous speech recognition on the domain of
broadcast news, the Switchboard database~\cite{switch92} that was designed for large-scale recognition for telephone conversions, and the AMI database~\cite{hain2007ami}
that consists of meeting recordings. These databases can be purchased from the institutes that created the data, but more generally from commercial organizations including
LDC\footnote{https://www.ldc.upenn.edu/} and ELRA\footnote{http://catalog.elra.info/}. For Chinese, the most popular database is the RAS 863 corpus~\cite{li2004rasc863},
which involves continuous reading speech of more than $80$ speakers, resulting in nearly $100$ hours of speech signals. Other commercial speech databases can be purchased from Datatang\footnote{http://www.datatang.com/}
and Speech Oceanf\footnote{http://www.speechocean.com/}.
The advantage of using these databases is that they are standard and publicly available, so that
results from different authors become comparable. This is very important for researchers to identify true promising algorithms and is actually one of
the most important factors that contribute to the recent success of the ASR research.

However, almost all these databases are not free; in fact, most of them are quite expensive. For example, the middle-sized WSJ corpus that was recorded $20$ years ago asks for $2,500$ US dollars\footnote{https://catalog.ldc.upenn.edu/LDC94S13A}. This licence fee is
affordable for most research institutes, however for individual researchers who just start their research activities or those who foster initial interests on
ASR, $2,500\$$ is rather expensive, particularly for researchers in developing countries. We can not complain the high licence fee because
organizing the recordings and annotation is very costly, and so it is not feasible to expect commercial institutes to provide free data.
Nevertheless, the high licence fee does form a barrier that blocks the initial interest to ASR of beginners. This is an intolerable situation
in the era where research activities have become more and more open, collaborative, and sharing.

\subsection{Free data movement}

We advocate a `free data' movement in the speech community. Data should not be private properties if they are collected by public institutes such as
universities and public-funded research groups. Since the collection and annotation is supported by public funds,  the data should be free for people who pay the fund, or even
for more broad people if the freedom will result in more benefit for the people who originally pay for the research. It is a shame that research projects
supported by public funds use money from taxpayers to build up private resources. Additionally, even commercial companies should consider release free data,
because the freedom will ultimately benefit the industrial sector: free data will
invoke more research in this area and so more opportunities to enjoy novel technic development and professional employees. Finally, the development of
internet accumulates large amount of speech data and the development of ASR technics allows cheaper annotation for these data. This makes the cost of
data collection is not as expensive as before, and thus free data is more feasible.

There have been some initial attempts towards this direction. For example, the EU-supported AMI/AMID project released all the data (both speech and video recordings on meetings),
the VoxForge project\footnote{http://www.voxforge.org/} provides a platform to publish GPL-based annotated speech audio.  OpenSLR is another platform to
publish open resources for speech and language research, including the LibriSpeech\footnote{http://www.openslr.org/12/}. The Sprakbanken database\footnote{http://www.nb.no/sbfil/talegjenkjenning/} is another free database in Swedish, Norwegian, Danish. For Chinese, we have never seen a free speech database
that is sufficient enough to build a full-fledged large-vocabulary Chinese ASR system.

\subsection{Our release}

To respect the creed that `public research should publish data' and break the data monopoly, to protect the initial interest of individual researchers towards
this research direction, the center for speech and language technologies (CSLT) at Tsinghua University recently started publishing free databases. The first database
we published is THUYG-20, which is used for training Uyghur speech recognition systems~\cite{rouze15}. This publication is collaborated with the Xinjiang university and the Tsinghua
AI Cloud Research Center (AICRC)\footnote{http://data.cslt.org/thuyg20/README.html}.

In this paper, we continue the free speech data process. We publish a Chinese database that was recorded by the first author $15$ years ago~\cite{wang01}. This database
consists of $35$ hours of speech signals recorded from $50$ participants. Associated with the speech signals, we provide the full set of resources including lexicon, LM and
the training recipe, so that new practitioners can use these resources to establish a basic large vocabulary continuous Chinese speech recognition system.  Interested
readers can download the data freely from \url{http://data.cslt.org/thchs30/README.html}.
To our best knowledge, this is the first release that can be used to build a practical Chinese speech recognition system. We hope this release can provide a standard reference for Chinese ASR researchers and invoke much enthusiasm of young people.

The rest of the paper is organized as follows: Section~\ref{sec:corpus} overviews the features of the released database, which we refer to as THCHS-30, and Section~\ref{sec:baseline} describes the baseline system we built based on the release data and resource. The paper is concluded in Section~\ref{sec:con}.

\section{Features of THCHS-30}
\label{sec:corpus}

We describe the THCHS-30 speech database in this section. This database was recorded in 2000 - 2001 by the first author when he was a master student, supervised by
Prof. Xiaoyan Zhu~\cite{wang01}. The design goal was a supplemental data resource for the 863 database, by which the phone coverage can be maximally improved.
The new database was named as TCMSD (Tsinghua Continuous Mandarin Speech Database), and has remains private usage since its creation.
We publish the data $15$ years later, with the permission from the licence holder, Prof. Zhu. The database was renamed as THCHS-30, standing for `Tsinghua Chinese 30 hour database'. The same nomenclature has been used for the THUYG-20 database, and will be followed by a number of free databases that we will publish soon.

\subsection{Speech signals}
\label{sec:design}

THCHS-30 involves more than $30$ hours of speech signals recorded by a single carbon microphone at the condition of silent office. Most of the participants are
young colleague students, and all are fluent in standard Mandarin. The sampling rate of the  recording is $16,000$ Hz, and the sample size is $16$ bits.

The sentences (text prompts in recording) of THCHS-30 were selected from a large volume of news wise, with the goal of augmenting the 863 database with more phone coverage. We selected $1000$ sentences for recording. Table~\ref{tab:cov} reports the bi-phone and tri-phone coverage with/without THCHS-30 (reproduced from~\cite{wang01}). It can be seen that THCHS-30 indeed improves phone coverage of the 863 database.

\begin{table}[!htbp]
\centering
\caption{Phone coverage with THCHS-30}
\label{tab:cov}
\centering
\begin{tabular}{l|c|c|c}
\hline
Database      & 863   & THCHS-30 & 863 + THCHS-30\\
\hline
No. Sentence     & 1500  & 1000     &  2500\\
Bi-phone Coverage &58.4\% & 71.5\%    & 73.4\%\\
Tri-phone coverage & 7.1\% & 14.3\%  & 16.8\% \\
\hline
\end{tabular}
\end{table}

The recordings are split into four groups according to the recording text: A (sentence ID from 1 to 250), B (sentence ID from 251 to 500), C (sentence ID form 501 to 750), D (sentence ID from 751 to 1000).
The utterances of group A, B and C are combined as the training set, which involves $30$ speakers and $10893$ utterances. All the utterances in group D are used as the test set, which involves $10$ speakers and $2496$ utterances. The statistics of THCHS-30 are represented in Table~\ref{tab:thchs}.

\begin{table*}[!htbp]
\centering
\caption{Statistics of THCHS-30 database}
\label{tab:thchs}
\centering
\begin{tabular}{l|c|c|c|c|c|c}
\hline
Data Set & Speaker& Male& Female& Age   & Utterance& Duration (hour)\\
\hline
Training & 30     & 8   & 22    & 20-55  & 10893    & 27.23h \\
Test     & 10     & 1   & 9     & 19-50  & 2496     & 6.24h  \\
\hline
\end{tabular}
\end{table*}

\subsection{Additional resources}

To assist constructing a practical Chinese ASR system, some additional resources are published in the THCHS-30 release. These resources include
lexica, language models, training recipes and some useful tools. Additional data in noise conditions are also provided.

\subsubsection{Lexicon and LM}

We release two language models and the associated lexica. The word-based LM involves 48k words and is based on word 3-grams, and the phone-based
LM involves $218$ Chinese tonal finial-initials and is based on phone 3-grams. The word LM was trained using a text collection that was
randomly selected from the Chinese Gigaword corpus\footnote{https://catalog.ldc.upenn.edu/LDC2003T09}. The training text involves $772,000$ sentences,
amounting to $18$ million words and $115$ million Chinese characters. The phone LM was trained using a much smaller text collection involving $2$ million characters.
The reason to use a smaller text data for phone LM training is that we want to retain linguistic information  as little as possible in the model
so that the resultant performance reflects more directly the quality of acoustic models.  The two LMs were trained with the SRILM tool~\cite{stolcke2002srilm}.

\subsubsection{Scripts and recipe}

The recipe and some utility scripts are also published for training a complete Chinese ASR system with THCHS-30. The scripts are based on the
framework of the Kaldi toolkit~\cite{Povey_ASRU2011}. The training process is basically similar to the wsj s5 GPU recipe provided by Kaldi, although some modifications
have been made to make it suitable for Chinese ASR.

\subsubsection{Noise data}

We are also interested in ASR tasks in noisy conditions, therefore provide a noisy version of THCHS-30: all the training and test data were corrupted by
three types of noise (separately): white noise, car noise and cafeteria noise. We focus on the $0$ db condition, which means that energy of noise and speech signals are
the same (thus very noisy). The noise embedding was performed by simple wave mixture, where the noise signals
are obtained from the public DEMAND noise repository\footnote{http://parole.loria.fr/DEMAND/}.

\subsection{Call for challenge}

Since THCHS-30 is public and free, everyone can download it, build their systems and compare their results with others. To encourage the research,
we call for challenge based on the provided resources,
including two tasks: large vocabulary recognition and phone recognition. Although the former is more close to practical usage, the latter can test
acoustic modeling approaches in a more focused way. For each task, we also compete for performance on the $0$ db noise condition with the three types of
noise: white, car and cafeteria.

\section{Baseline system}
\label{sec:baseline}

We describe our baseline system built with THCHS-30, and report the performance in various conditions that we have called for challenge.
We treat these results as the baseline for the competition; any improvement by any authors will be published on the competition web page\footnote{http://data.cslt.org/thchs30/challenges/asr.html}.

\subsection{Framework and setting}

\begin{figure}[htb]
\begin{minipage}[b]{1.0\linewidth}
  \centering
  \centerline{\includegraphics[width=8.0cm]{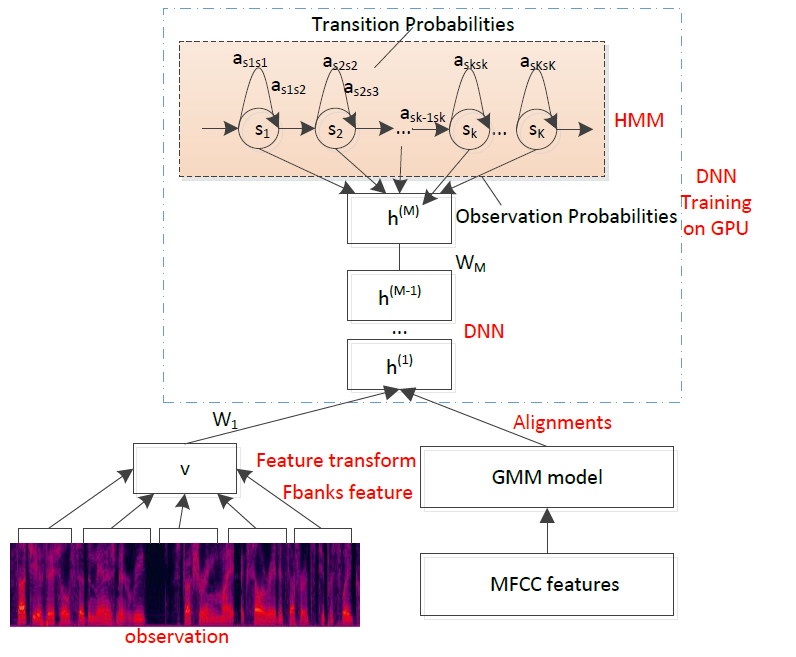}}
\end{minipage}
\caption{Architecture of the THCHS-30 baseline system.}
\label{fig:arc}
\end{figure}

We use the Kaldi toolkit to train the HMM-DNN hybrid acoustic models. The architecture of the system is shown in Fig~\ref{fig:arc}. Following this architecture, a monophone GMM system was firstly trained with the standard $13$ dimensional MFCCs plus the first and second order derivatives. The cepstral mean normalization (CMN) was employed to reduce the channel effect. A triphone GMM system was then constructed based on the monophone system with features transformed by LDA and MLLT. The final GMM system was used to generate state alignment for the subsequent DNN training~\cite{Sequence}.

The DNN system was trained based on the alignments provided by the GMM system. The features were $40$-dimensional FBanks, and neighbouring frames were concatenated by a 11-frame window (5 windows on each side). The concatenated features were transformed by LDA, by which the dimensionality was reduced to 200. A global mean and variance normalization was then applied to obtain the DNN input. The DNN architecture consists of $4$ hidden layers and each layer consists of $1200$ units. The output layer is composed of $3386$ units. The baseline DNN model was trained with the criterion of cross entropy. The stochastic gradient descendent (SGD) algorithm was used to performance the optimization. The mini batch size was set to $256$ frames, and the initial learning rate was set to be 0.008.

\subsection{Preliminary results}

The performance of the DNN system based on clean training data is presented in Table~\ref{tab:base}, where the word ASR task is evaluated in terms of character error rate (CER) and the phone ASR task is evaluated in terms of phone error rate (PER). Table~\ref{tab:base} reports the test results on both the clean speech and the $0$ db noisy
speech. It can be seen that the performance  on noisy speech is much lower than that on clean speech, particularly when the corruption is white noise. Note that the CER and PER are rather high compared to those obtained with many standard databases (e.g., 863 database), even on clean speech. This is mostly attributed to the `abnormality' of the test data: to seek for maximal phone coverage, the sentences selected for THCHS-30 tend to be rather peculiar in both pronunciation and spelling. This makes the recognition task highly challenging, and solving the peculiarity perhaps needs some specific techniques for example more aggressive phone sharing.

\begin{table}[!htbp]
\centering
\caption{Performance of THCHS-30 baseline system trained with clean speech. }
\label{tab:base}
\centering
\begin{tabular}{l|c|c}
\hline
Condition        &  Word Task (CER\%) & Phone Task (PER\%)\\
\hline
GMM Clean            & 36.91 & 20.21 \\
DNN Clean            & 30.11  & 14.81\\
\hline
$0$db-White      & 96.44  & 71.24\\
$0$db-Car        & 34.22  & 18.16\\
$0$db-Cafeteria  & 79.83  & 54.03\\
\hline
\end{tabular}
\end{table}

\subsection{Noise cancellation with DAE}

The significant performance degradation on noisy data can be addressed in multiple ways, for example, multiple condition training~\cite{dnn:yu13:288}
and noisy training~\cite{dnn:chinasip14,yin2015noisy}. Although promising, these approaches require re-training the DNN model and so is
not convenient for dealing with new noise types. This paper applies a noise cancellation approach based on deep auto-encoder (DAE).
DAE is a special implementation of autoencoder (AE), by introducing random corruptions
to the input features in model training. It has been shown that this model is very powerful
in learning low-dimensional representations and can be used to recover noise-corrupted input~\cite{vincent2008extracting}.
In~\cite{maas2012recurrent}, DAE is extended to a deep recurrent structure and has been employed to recover clean speech
in noisy conditions for ASR. A recent study employs DAE in de-reverberation~\cite{ishii2013reverberant}, and our work also showed that
it can be used for removing music corruption~\cite{zhaomusic}.

We employ a simple DAE for noise cancellation in this study. Firstly noisy data are generated by corrupting clean speech with noise signals, and then a
non-linear transform function can be learned with these data. Since the noisy data are generated artificially, the only cost is to find a piece of noise signal
and use it to corrupt clean speech. The noisy data generation process follows a scheme similar to the one in~\cite{yin2015noisy}, where a particular SNR is
sampled for each clean utterance, following a Gaussian distribution whose mean is $0$ db. The sampled SNR is then used as the corruption level to generate
the noisy version of the utterance. The training process of the DAE model is the same as the one used to train the  DNN acoustic model, except that the
objective function is replaced as reconstruction error.

In practice, DAE is used as a particular component of the front-end pipeline. The input involves a window of $11$ frames Fbank features (after mean normalization),
and the output is the noise-removed feature corresponding to the central frame. The outputs are then treated as regular Fbank features that are spliced, LDA transformed and
globally normalized before fed into the DNN acoustic model (trained with clean speech).

The CER/PER results of the DNN baseline system on noisy data with the DAE-based noise cancellation are reported in Table~\ref{tab:dae}. Compared to the numbers in
Table~\ref{tab:base}, it can be observed that the performance is significantly improved after noise removed by DAE. This approach is flexible to address any noise
by learning a noise-specific DAE, only with a small piece of noise example.

\begin{table}[!htbp]
\centering
\caption{Performance of THCHS-30 baseline system  with DAE-based noise cancelation.}
\label{tab:dae}
\centering
\begin{tabular}{l|c|c}
\hline
Condition        &  Word Task (CER\%) & Phone Task (PER\%)\\
\hline
$0$db-White      & 75.01  & 46.95\\
$0$db-Car        & 32.13  & 15.96\\
$0$db-Cafeteria  & 56.37  & 32.56\\
\hline
\end{tabular}
\end{table}

\section{Conclusions}
\label{sec:con}

The aim of the paper is to publish a free 30-hour Chinese speech database, THCHS-30. Additional resources such as  lexica, LMs and training recipes are also published to assistant new researchers building their first ASR systems. We demonstrated the building process and presented the baseline results on both clean and noisy data.

As far as we know, this is the first free Chinese speech database that can be used to build a practical Chinese ASR system. We hope this release will attract more potential researchers to this promising field, particularly young researchers whose initial interest would otherwise killed by the high licence fee of commercial databases. We call for challenges based on this database, and hope that invokes deep innovation and more collaboration.

\section*{Acknowledgement}

This research was supported by the National Science Foundation of China (NSFC) under the project No. 61371136, and the MESTDC PhD Foundation Project No. 20130002120011. It was also supported by Sinovoice and Huilan Ltd.

%\vfill\pagebreak

\bibliographystyle{IEEEbib}
\bibliography{refs}

\begin{thebibliography}{10}

\bibitem{RECENT}
Li~Deng, Jinyu Li, Jui-Ting Huang, Kaisheng Yao, Dong Yu, Frank Seide,
  Michael~L. Seltzer, Geoff Zweig, Xiaodong He, Jason Williams, Yifan Gong, and
  Alex Acero,
\newblock ``Recent advances in deep learning for speech research at
  microsoft,''
\newblock {\em Acoustics, Speech and Signal Processing (ICASSP), 2013 IEEE
  International Conference on}, 2013.

\bibitem{Perspective}
Xuedong Huang, James Baker, and Raj Reddy,
\newblock ``A historical perspective of speech recognition,''
\newblock {\em Communications of the ACM}, vol. 57, no. 1, pp. 94--103, 2014.

\bibitem{automatic}
Dong Yu and Li~Deng,
\newblock {\em Automatic Speech Recognition A Deep Learning Approach},
\newblock Springer, 2015.

\bibitem{Deep}
Dong Yu and Li~Deng,
\newblock {\em Deep Learning Methods and Applications},
\newblock Foundations and Trends in Signal Processing, 2014.

\bibitem{Fast}
Chenghao Cai, Yanyan Xu, Dengfeng Ke, and Kaile Su,
\newblock ``A fast learning method for multilayer perceptrons in automatic
  speech recognition systems,''
\newblock {\em Journal of Robotics}, vol. 2015, no. 2015, pp. 1--7, 2015.

\bibitem{Discriminative}
Xiaodong He, Li~Deng, and Wu~Chou,
\newblock ``Discriminative learning in sequential pattern recognition,''
\newblock {\em IEEE SIGNAL PROCESSING MAGAZINE}, pp. 14--36, 2008.

\bibitem{Phoneme}
Carla Lopes and Fernando Perdigo,
\newblock {\em Phoneme Recognition on the TIMIT Database},
\newblock Speech Technologies, 2011.

\bibitem{Design}
D.~Paul and J.~Baker,
\newblock ``The design of wall street journal-based csr corpus,''
\newblock {\em Proceedings of the International Conference on Spoken Language
  Systems (ICSLP)}, pp. 899--902, 1992.

\bibitem{switch92}
J.J. Godfrey, E.C. Holliman, and J.~McDaniel,
\newblock ``Switchboard: telephone speech corpus for research and
  development,''
\newblock in {\em Acoustics, Speech, and Signal Processing, 1992. ICASSP-92.,
  1992 IEEE International Conference on}, Mar 1992, vol.~1, pp. 517--520 vol.1.

\bibitem{hain2007ami}
Thomas Hain, Vincent Wan, Lukas Burget, Martin Karafiat, John Dines, Jithendra
  Vepa, Giulia Garau, and Mike Lincoln,
\newblock ``The ami system for the transcription of speech in meetings,''
\newblock in {\em Acoustics, Speech and Signal Processing, 2007. ICASSP 2007.
  IEEE International Conference on}. IEEE, 2007, vol.~4, pp. IV--357.

\bibitem{li2004rasc863}
Aijun Li, Zhigang Yin, Tianqing Wang, Qiang Fang, and Fang Hu,
\newblock ``Rasc863-a chinese speech corpus with four regional accents,''
\newblock {\em ICSLT-o-COCOSDA, New Delhi, India}, 2004.

\bibitem{rouze15}
Askar Rouze, Shi Yin, Zhiyong Zhang, Dong Wang, Askar Humdulla, and Fang Zheng,
\newblock ``Thugy20: A free uyghur speech database,''
\newblock in {\em NCMMSC 2015}, 2015.

\bibitem{wang01}
Dong Wang, Dalei Wu, and Xiaoyan Zhu,
\newblock ``{TCMSD:} a new chinese continuous speech database,''
\newblock in {\em International Conference on Chinese Computing (ICCC’01),
  2001,}, 2001.

\bibitem{stolcke2002srilm}
Andreas Stolcke,
\newblock ``{SRILM}-an extensible language modeling toolkit,''
\newblock in {\em ICSLP2002}, 2002.

\bibitem{Povey_ASRU2011}
Daniel Povey, Arnab Ghoshal, Gilles Boulianne, Lukas Burget, Ondrej Glembek,
  Nagendra Goel, Mirko Hannemann, Petr Motlicek, Yanmin Qian, Petr Schwarz, Jan
  Silovsky, Georg Stemmer, and Karel Vesely,
\newblock ``The kaldi speech recognition toolkit,''
\newblock in {\em IEEE 2011 Workshop on Automatic Speech Recognition and
  Understanding}. Dec. 2011, IEEE Signal Processing Society,
\newblock IEEE Catalog No.: CFP11SRW-USB.

\bibitem{Sequence}
K.~Vesely, A.~Ghosal, L.~Burget, and D.~Povey,
\newblock ``Sequence-discriminative training of deep neural networks,''
\newblock {\em Proceedings of the 14th Annual Conference of the International
  Speech Communication Association}, 2013.

\bibitem{dnn:yu13:288}
Dong Yu, Michael~L Seltzer, Jinyu Li, Jui-Ting Huang, and Frank Seide,
\newblock ``Feature learning in deep neural networks - a study on speech
  recognition tasks,''
\newblock in {\em Proc. of International Conference on Learning
  Representations}, 2013.

\bibitem{dnn:chinasip14}
Xiangtao Meng, Chao Liu, Zhiyong Zhang, and Dong Wang,
\newblock ``Noisy training for deep neural networks,''
\newblock in {\em Proc. of ChinaSIP 2014}, 2014, pp. 16--20.

\bibitem{yin2015noisy}
Shi Yin, Chao Liu, Zhiyong Zhang, Yiye Lin, Dong Wang, Javier Tejedor,
  Thomas~Fang Zheng, and Yinguo Li,
\newblock ``Noisy training for deep neural networks in speech recognition,''
\newblock {\em EURASIP Journal on Audio, Speech, and Music Processing}, vol.
  2015, no. 1, pp. 1--14, 2015.

\bibitem{vincent2008extracting}
Pascal Vincent, Hugo Larochelle, Yoshua Bengio, and Pierre-Antoine Manzagol,
\newblock ``Extracting and composing robust features with denoising
  autoencoders,''
\newblock in {\em Proceedings of the 25th international conference on Machine
  learning}. ACM, 2008, pp. 1096--1103.

\bibitem{maas2012recurrent}
Andrew~L Maas, Quoc~V Le, Tyler~M O'Neil, Oriol Vinyals, Patrick Nguyen, and
  Andrew~Y Ng,
\newblock ``Recurrent neural networks for noise reduction in robust asr.,''
\newblock in {\em INTERSPEECH}. Citeseer, 2012.

\bibitem{ishii2013reverberant}
Takaaki Ishii, Hiroki Komiyama, Takahiro Shinozaki, Yasuo Horiuchi, and Shingo
  Kuroiwa,
\newblock ``Reverberant speech recognition based on denoising autoencoder.,''
\newblock in {\em INTERSPEECH}, 2013, pp. 3512--3516.

\bibitem{zhaomusic}
Mengyuan Zhao, Dong Wang, Zhiyong Zhang, and Xuewei Zhang,
\newblock ``Music removal by denoising autoencoder in speech recognition,''
\newblock in {\em APSIPA2015}, 2015.

\end{thebibliography}

\end{document}